\documentclass[letterpaper, 10 pt, conference]{ieeeconf}  
\IEEEoverridecommandlockouts 

\usepackage{graphics}
\usepackage{epsfig}
\usepackage{mathptmx}
\usepackage{times}
\usepackage{amsmath}
\usepackage{algorithm}
\usepackage[noend]{algpseudocode}
\usepackage{lineno}
\usepackage{amssymb}
\usepackage{hyperref}
\usepackage{balance}
\usepackage{multirow}
\usepackage{cite}
\usepackage{paralist}
\usepackage{bm}
\let\labelindent\relax
\usepackage{enumitem}
\usepackage{comment}
\usepackage{microtype}

\title{\LARGE \bf
Automatic Interaction and Activity Recognition from Videos of \\ Human Manual Demonstrations with Application to Anomaly Detection
}

\author{Elena Merlo$^{1,2}$, Marta Lagomarsino$^{1,3}$, Edoardo Lamon$^{1,4}$, and Arash Ajoudani$^{1}$
\thanks{This work was supported by the ERC-StG Ergo-Lean (Grant Agreement No. 850932) and in part by the European Union under NextGenerationEU (FAIR - Future AI Research - PE00000013).}
\thanks{$^1$ Human-Robot Interfaces and Interaction Laboratory, Istituto Italiano di Tecnologia, Genoa, Italy. \tt\small elena.merlo@iit.it}
\thanks{$^2$ Dept. of Informatics, Bioengineering, Robotics, and Systems Engineering, University of Genoa, Genoa, Italy.}
\thanks{$^3$ Dept. of Electronics, Information and Bioengineering, Politecnico di Milano, Milan, Italy.}
\thanks{$^4$ Dept. of Information Engineering and Computer Science, University of Trento, Trento, Italy.}
}

\begin{document}

\maketitle
\thispagestyle{empty}
\pagestyle{empty}

\begin{abstract}
This paper presents a new method to describe spatio-temporal relations between objects and hands, to recognize both interactions and activities within video demonstrations of manual tasks. The approach exploits Scene Graphs to extract key interaction features from image sequences while simultaneously encoding motion patterns and context. 
Additionally, the method introduces event-based automatic video segmentation and clustering, which allow for the grouping of similar events and detect if a monitored activity is executed correctly.
The effectiveness of the approach was demonstrated in two multi-subject experiments, showing the ability to recognize and cluster hand-object and object-object interactions without prior knowledge of the activity, as well as matching the same activity performed by different subjects.
\end{abstract}

\section{INTRODUCTION}
\label{sec:introduction}

The comprehension of human activities enables machines to understand and interpret the visual information they acquire and make informed decisions based on what they see~\cite{weinland2011survey}.
Many robotic applications rely on video comprehension, such as autonomous navigation, interactions with objects, and human-robot interaction. In particular, in the latter, it is of paramount importance to be able to recognize human activities and predict their outcomes, for instance in scenarios where robots provide assistance in Activities of Daily Living (ADL)~\cite{massardi2020parc} or collaborate with humans in industrial settings to achieve a common task~\cite{lagomarsino2022pick}.
Moreover, by understanding an activity demonstrated by humans, robots could learn the task structure and replicate it~\cite{ramirezamaro2017transferring}.
Both domestic and industrial activities are characterized by the prominent presence of manual tasks.
However, being able to transfer detailed knowledge about human manipulation activities is a challenging problem. When addressing such a challenge, a question arises: "What do we want the robot to learn?", and two are the potential answers~\cite{ramirezamaro2017transferring}. The first resides in teaching the robot to reproduce exact movements, at the trajectory level \cite{albrecht2011imitating}, while the second is to replicate the outcome of the activity, understanding the activity goal and context, not only in physical but also in semantic terms \cite{patterson2005fine, aksoy2010categorizing, wachter2013action, guha2013minimalist}.
The first approach has been extensively studied within the Learning by Demonstration paradigm and in general requires accurate motion capture tools which might not always be available, as in the case the activity demonstration consists of pre-recorded videos.
For these reasons, the manuscript will focus on the second approach.

A pioneering study in this direction is \cite{kuniyoshi1994learning}, which leverages the concept of active vision to comprehend the content of a series of images, extracting regions of interest and specific descriptive features.
In our framework, the feature selection has been carefully conducted to enable the description of activities and facilitate their replication by the robot. Since we are dealing with manual activities, the primary interest lies in describing hand-object and object-object interactions. Our key features, also employed in the cited works, include the spatial location of hands and objects and their relative distances.
Moreover, we enclose more detailed information about movements during an interaction, such as the velocity and direction of hands and objects involved~\cite{fu2002temporal}.

To provide a more detailed understanding of the interaction flows, we propose a method to include all relevant features through a scene encoder and to automatically segment activities hierarchically, according to the definition of our taxonomy.
Then, we process the segmented time series of features to capture the temporal patterns of the interactions and recognize similar ones.
The method we propose adopts an unsupervised approach, which grants flexibility to describe various types of interactions, avoids the challenge associated with obtaining labeled data, and opens possibilities for discovering hidden patterns and structures within the data.
By leveraging the intrinsic characteristics of interactions, our goal is to capture the underlying similarities and differences among them, enabling the recognition based on such properties rather than relying on any prior knowledge~\cite{kim2021motion, ahn2021refining}.
Since our encoding can separate contextual information from motion information, it enables us to recognize, as similar interactions, analogous motion patterns with different objects and, at the same time, distinguish different motion patterns executed with the same objects. For example, assembling and disassembling two pieces might involve the same objects but opposite motions.
Finally, individual activities are recognized by identifying time sequences of frames corresponding to specific interactions. In this way, by comparing the execution of an activity with the learned ones through the proposed metrics, the method could also be used to detect potential anomalies in the activity's progress.

The method has been tested in two multi-subject experiments (see multimedia attachment and \href{https://youtu.be/Ftu_EHAtH4k}{youtu.be/Ftu\_EHAtH4k})) the capability of the method in finding similarities in activities executed by different subjects and ii) the potential of the method in detecting in-progress activity anomalies given a learned activity representation.
The results demonstrate the effectiveness of the proposed approach in recognizing and clustering hand-object and object-object interactions and in matching the same activity executed by different subjects.

\section{EVENT-BASED TAXONOMY}
\label{sec:taxonomy}
In this study, we focus on the analysis of manual activities that involve the manipulation of objects through human hands. Henceforth, we will refer to both objects and human hands as \textit{video objects}, which are, using terminology introduced in \cite{fu2002temporal}, specific entities or regions of interest within a video sequence that a computer system is able to identify and track over time.

In our approach, we examine the spatio-temporal relationships between video objects by analyzing the content of smaller video segments obtained by video segmentation. By segmenting the video into atomic units and then grouping these interactions according to logic rules, it is possible to describe and recognize the activity occurring in the video. In other words, we are aiming to understand higher-level activities starting from descriptions of low-level interactions, adopting a bottom-up hierarchical approach. To further clarify the problem and the goal of the work, we provide an event-based taxonomy, which extends the one proposed by \textit{Fu et al.}~\cite{fu2002temporal} to define each different event:

\begin{itemize}
    \item an \textbf{Elementary Reaction Unit} (ERU) is a set of consecutive frames within which video objects have a specific spatio-temporal relationship.  
    The onset of a new ERU is due to a change in the video objects' relationship;
    \item an \textbf{Interaction Unit} (IU) is a time-ordered sequence of ERUs that involve the same video objects. By grouping the ERUs, we are able to capture all of the changes that occur in the spatio-temporal relationship between the video objects;
    \item an \textbf{activity} is a time-ordered sequence of IUs. The IUs within an activity are logically connected, with the successful completion of one IU leading to the start of the next. When two IUs are not connected logically it means that the activity is changed;
    \item a \textbf{job} is a collection of activities which share an overall common objective.
\end{itemize}
In particular, the definitions of ERU and IU are based on \cite{fu2002temporal} (see also \autoref{fig:segmentation}). 
However, the taxonomy lacks the ability to capture the overall structure and context of the interactions. Therefore, we introduced the activity and job layers, which allow us to group hierarchically IUs into more general, human-understandable units.
\section{METHODOLOGY}
\label{sec:methodology}

\begin{figure*}[!t]
    \centering
	\includegraphics[trim=0.0cm 0.0cm 2.1cm 0.0cm,clip,width=1\linewidth]{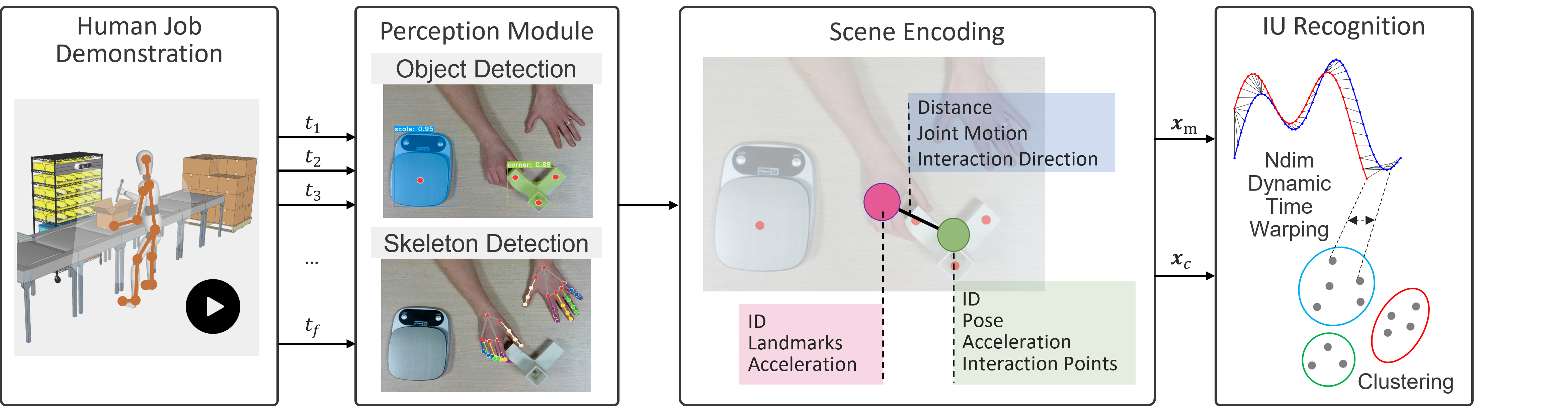}
    \caption{Overview of the overall framework. Given a video demonstration of a human job, the perception module detects and extracts the 3D objects poses and hands landmarks positions. In red, the interaction points for each object. This data is organized per frame into a Scene Graph structure, where objects and hands are the nodes, and edges represent interactions between the connected nodes. The scene's content is encoded into a feature space, capturing both motion $\boldsymbol{x}_\text{m}$ and context $\boldsymbol{x}_\text{c}$ at the interaction level. Changes in the feature space indicate variations in the scene, leading to the segmentation of the video into Interaction Units (IU)s. IUs can be recognized and compared using clustering techniques to group together similar ones.}
	\label{fig:action_recognition_scheme}
	\vspace{-0.5cm}
\end{figure*}

The framework depicted in \autoref{fig:action_recognition_scheme} illustrates the various stages of the proposed approach. The input of the framework is a video demonstration of the manipulation task. The perception module is responsible for extracting the 3D hands' landmarks and the 3D pose of the objects in the scene.
It should be noted that the presented method is not limited to a particular technique to detect hand and object movements. Instead, it can accommodate various approaches, including inertial motion capture, marker-based motion capture with multiple cameras, or RGBD-based markerless motion capture. 
Additionally, information on the objects type and interaction points are required as input. These points are chosen based on the object properties such as shape, size, and affordance, and correspond to the suitable grasp frames for object interactions with hands or other objects~\cite{ardon2019reasoning}.
Similarly, providing a technique to automatically retrieve the object interaction points is out of the scope of the manuscript. However, SoA techniques in robotic manipulation and grasping can be employed~\cite{carbone2013grasping}.
Once extracted, the perception module outputs are organized per frame into a Scene Graph (SG) structure \cite{johnson2015image, chang2023comprehensive} that portrays the spatial semantic relationships among the video objects in the scene. This representation, enriched with local temporal features, such as video objects' relative accelerations and velocities, could provide a detailed and sensitive-to-variation scene description.
In this way, specific changes in the feature space represent variations in the scene content, leading to the segmentation of the video into ERUs, IUs, or activities. 
Once the video has been segmented, it is possible to recognize and compare different patterns. In particular, in this manuscript, we will focus on grouping IUs using clustering techniques, such as centroid- or distribution-based clustering~\cite{arthur2007kmeans, schubert2017dbscan}.
This involves comparing the IUs based on similarity measurements and grouping together the similar ones. The feature set consists of two components. The first component encodes motion features at the interaction level, regardless of which objects are involved. The second one encodes contextual information, such as the identity of the objects involved in the interaction.
When clustering motion features, we obtain clusters of context-free IUs that are characterized by similar motion patterns. Instead, when clustering contextual information, we group IUs that involve the same objects being handled. The latter allows us to identify interactions based on context that are equal in terms of the specific objects involved, even if their motion features differ.
However, to discriminate IUs, both feature components are necessary. Therefore, an ensemble-like approach is used, where the results of each independent clustering are combined into a single cluster.

\subsection{Scene Encoding}
By processing the perception data, each frame representing a scene is mapped into a SG. 
A SG $G$ is defined as the tuple $G = (V, R, E)$. $V = \{v_\text{1}, \dots , v_\text{$|\text{v}|$}\}$ is the set of video objects which are interacting with each other (foreground). We discard not interacting objects since they are considered not relevant in the scene description (background). Each object is represented as $v_\text{i} = (c_\text{i}, \hspace{1mm} A_\text{i})$, where $c_\text{i}$ and $A_\text{i}$ respectively indicate the category and attributes of the video object. $R$ represents a set of relationships between the nodes, while $E$ denotes the edges between the two interacting video objects~\cite{chang2023comprehensive}. Furthermore, in order to describe the interactions for each hand separately as in~\cite{dreher2020learning}, we will generate $n$ SGs if there are $n$ hands interacting with the same object. This allows us to capture the unique interactions of each hand with the objects.

\subsubsection{Scene Graph nodes}
Each node $v \in V$ in a scene graph represents a video object, with different attributes depending on the type. 
The $i$-$th$ hand node is represented by $v_{\text{h}_\text{i}} = (c_{\text{h}_\text{i}}, \hspace{1mm} A_{\text{h}_\text{i}})$, with $c_{\text{h}_\text{i}} = ID_{\text{h}_\text{i}}$, and $A_{\text{h}_\text{i}} = (LM_{\text{h}_\text{i}}, \hspace{1mm} \alpha_{\text{h}_\text{i}})$, where $ID_\text{h}$ is the hand identity, $\alpha_{\text{h}_\text{i}}$ denotes the tangential component of the hand acceleration measured at the middle finger knuckle, and $LM_{\text{h}_\text{i}}$ is the set of hand landmarks which represent key locations such as fingertips, knuckles, and wrist.  
A subset of $LM_{\text{h}_\text{i}}$, in particular the landmarks corresponding to the fingertips of the three middle fingers, represents the hand interaction points $IP_{\text{h}_\text{i}}$. 
Instead, the $j$-$th$ object node is represented by $v_{\text{o}_\text{j}} = (c_{\text{o}_\text{j}}, \hspace{1mm} A_{\text{o}_\text{j}})$, with $c_{\text{o}_\text{j}} = ID_{\text{o}_\text{j}}$, and $A_{\text{o}_\text{j}} = (\boldsymbol{p}_{\text{o}_\text{j}}, \hspace{1mm} \boldsymbol{\phi_{\text{o}_\text{j}}}, \hspace{1mm} \alpha_{\text{o}_\text{j}}, \hspace{1mm} IP_{\text{o}_\text{j}})$, where $ID_{\text{o}_\text{j}}$ is the object identity, $\boldsymbol{p}_{\text{o}_\text{j}}$ is the position (based on the object centroid), $\boldsymbol{\phi}_{\text{o}_\text{j}}$ is the orientation, $\alpha_{\text{o}_\text{j}}$ denotes the tangential component of the object acceleration measured at the object centroid, and $IP_{\text{o}_\text{j}}$ is the set of the interaction points position, computed with respect to $\boldsymbol{p}_{\text{o}_\text{j}}$.

\subsubsection{Scene Graph edges}
An edge $e_\text{i,j} \in E$ connects two video objects $v_\text{i}$ and $v_\text{j}$ if and only if
\begin{equation*}
    d_\text{i,j} = dist(IP_{v_\text{i}}^\text{ k}, \hspace{1mm} IP_{v_\text{j}}^\text{ q}) \approx 0, \quad \forall \; k, q \in [1, \hspace{1mm} |IP_{v}|], \; k \neq q,
\end{equation*}  
where $dist(\cdot, \cdot)$ is the Euclidean distance. In other words, we determine whether there is an interaction between two video objects if the minimum Euclidean distance between each pair of their respective interaction points $IP_{v_\text{i}}$ and $IP_{v_\text{j}}$ is reasonably small as in \cite{ramirezamaro2017transferring}. 
Each $e_\text{i,j} \in E$ is described by a relationship between the connected nodes $v_\text{i}$ and $v_\text{j}$, $r_\text{i $\rightarrow$ j} \in R$. 
Each $r_\text{i $\rightarrow$ j} \in R$ includes three attributes:
\begin{itemize}
    \item the \textit{distance} $d_\text{i,j}$ between the two objects $v_\text{i}$ and $v_\text{j}$;
    \item the \textit{joint motion}: whether $v_\text{i}$ and $v_\text{j}$ are moving jointly;
    \item the \textit{relative motion direction}, expressing the direction along which $v_\text{i}$ is moving towards or away from $v_\text{j}$.
\end{itemize}

To determine whether the two sufficiently close video objects are moving jointly, we compare their tangential acceleration components. 
If $sgn(\alpha_{\text{v}_\text{i}})$ and $sgn(\alpha_{\text{v}_\text{j}})$ are concordant, we can conclude that $v_\text{i}$ and $v_\text{j}$ are moving together. In the case where a hand and an object are jointly moving, we obtain the \textit{in-hand} relation between them, i.e. we assume that the hand holds the object as in\cite{ramirezamaro2014automatic}. 
We expand the concept of the in-hand relationship beyond just a hand and an object, to determine if multiple objects function as an integral unit, such as in the case of an assembly.

In addition, the interaction direction is obtained by projecting the velocity vector of $v_\text{i}$ onto the frame $T_{v_\text{j}} = [\boldsymbol{p}_{v_\text{j}}, \boldsymbol{\phi_{v_\text{j}}}]$ of the approached/left video object $v_\text{j}$; the resulting vector is transformed into spherical coordinates, i.e., elevation  $\theta_\text{i,j}$ and azimuth $\varphi_\text{i,j}$ angles, and radius $\rho_\text{i,j}$. These coordinates are then quantized:
$\theta_\text{i,j}^\text{Q} = Q(\theta_\text{i,j}), \quad\varphi_\text{i,j}^\text{Q} = Q(\varphi_\text{i,j}).$
As a result, $\theta_\text{i,j}^\text{Q}$ and $\varphi\text{i,j}^\text{Q}$ describe the interaction direction of video objects using a finite number of integer values.
Retrieving the interaction direction helps to distinguish activities of the same type, differentiated only by the direction in which the objects interact with each other. This feature is also useful to recognize if the interaction is well performed or warped. This representation goes beyond the simple away/toward dichotomy and encompasses additional dimensions such as right/left, bottom/up, in front/behind, and all their possible combinations \cite{fu2002temporal}. Moreover, another robust point of our design is the space invariance, by encoding in the same way interactions that involve similar relative approaching directions regardless of the video objects' absolute poses. 

\subsubsection{Feature Couple}
The scene representation provided by the SG can be further reduced by means of a feature couple, denoted as $X=(\boldsymbol{x}_m, \boldsymbol{x}_c)$, where $\boldsymbol{x}_\text{m}$ conveys semantic motion information (\textit{motion features}), while $\boldsymbol{x}_\text{c}$ about the video objects IDs (\textit{context features}).
To generate the two feature vectors, we employ a Dijkstra search algorithm \cite{dijkstra2022note} on the SG, utilizing the distance edge attribute as the cost metric. This process enables us to identify and consider the hand-object and object-object interactions that exist along the optimal path in the graph, starting from the hand node. By traversing nodes and edges on this path, features are extracted and stored in the $\boldsymbol{x}_m$ and $\boldsymbol{x}_c$ vectors. In the case of a hand-object interaction between $v_{\text{h}_\text{i}}$ and $v_{\text{o}_\text{j}}$, and an object-object interaction between $v_{\text{o}_\text{j}}$ and $v_{\text{o}_\text{l}}$, the two vectors will have this form: 
\begin{align*}
   \boldsymbol{x}_\text{m} & = \begin{bmatrix}  a_{\text{h}_\text{i}} & \theta_{\text{h}_\text{i}, \text{o}_\text{j}}^\text{Q} & \varphi_{\text{h}_\text{i}, \text{o}_\text{j}}^\text{Q} & jm_{\text{h}_\text{i}, \text{o}_\text{j}} & \theta_{\text{o}_\text{j}, \text{o}_\text{l}}^\text{Q} & \varphi_{\text{o}_\text{j}, \text{o}_\text{l}}^\text{Q} & jm_{\text{o}_\text{j}, \text{o}_\text{l}} \end{bmatrix} \\
   \boldsymbol{x}_\text{c} & = \begin{bmatrix}  ID_{\text{h}_\text{i}} & ID_{\text{o}_\text{j}} & ID_{\text{$\text{o}_\text{j}$}} & ID_{\text{$\text{o}_\text{l}$}} \end{bmatrix}
\end{align*}
In $\boldsymbol{x}_\text{m}$, the first feature $a_{\text{h}_\text{i}} = sgn(\alpha_{\text{h}_\text{i}})$ represents the hand acceleration categorized into three states: accelerating, decelerating, or maintaining a constant velocity (if $\alpha_{\text{h}_\text{i}} \approx 0$).
The elements $\theta_{\text{h}_\text{i}, \text{o}_\text{j}}^\text{Q}$, $\varphi_{\text{h}_\text{i}, \text{o}_\text{j}}^\text{Q}$, $\theta_{\text{o}_\text{j}, \text{o}_\text{l}}^\text{Q}$, $\varphi_{\text{o}_\text{j}, \text{o}_\text{l}}^\text{Q}$ represent the direction of the interactions, while $jm_{\text{h}_\text{i}, \text{o}_\text{j}}$, $jm_{\text{o}_\text{j}, \text{o}_\text{l}}$ encode the joint motion, i.e. whether the interacting video objects are integral. Finally, $\boldsymbol{x}_\text{c}$ contains their IDs.
Note that all values in the feature vectors are integers, and each element in $\boldsymbol{x}_\text{m}$ belongs to an ordered range with a unit distance between consecutive values.

\subsection{Event-Based Video Segmentation}
\label{subsec:segmentation}
\begin{figure}[!t]
    \centering
	\includegraphics[trim=0.0cm 0.0cm 0.0cm 0.0cm,clip,width=0.9\linewidth]{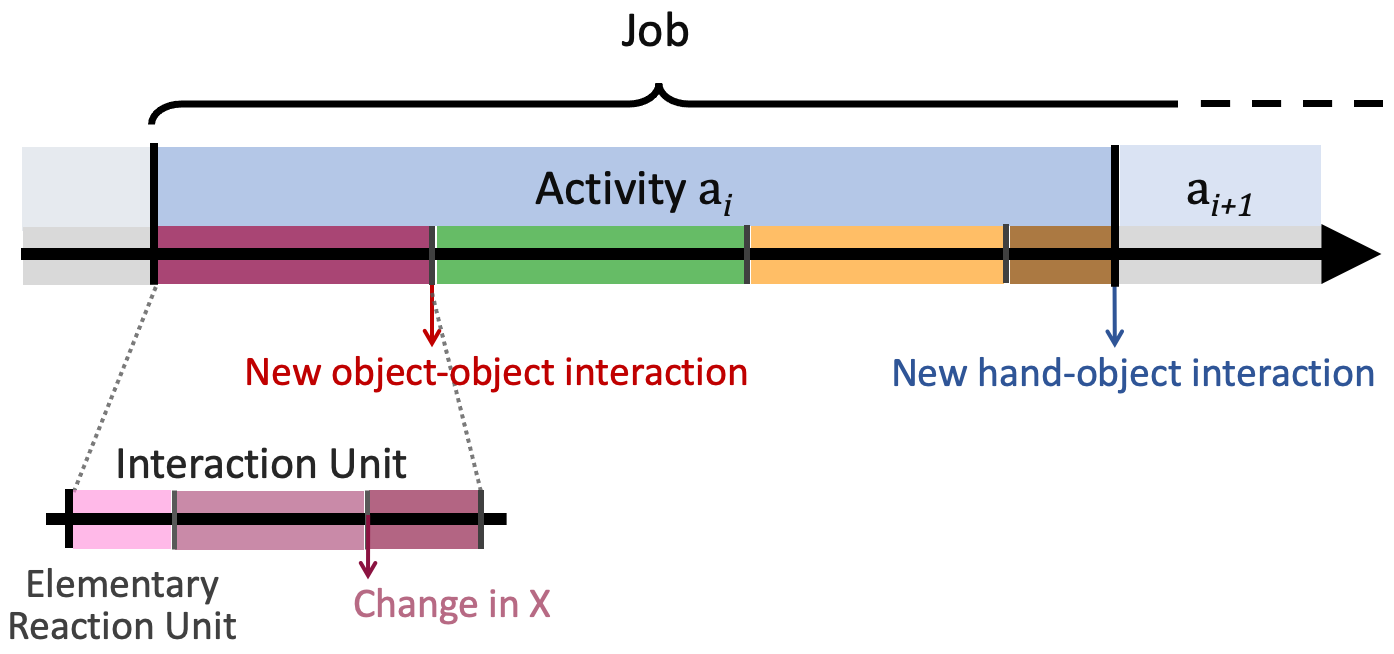}
    \vspace{-0.3cm}
    \caption{Conceptual illustration of the proposed taxonomy and automatic video segmentation criteria.}
	\label{fig:segmentation}
	\vspace{-0.5cm}
\end{figure}

Using the proposed encoding from a time series of video frames, we obtain a time series of $X$. The video segmentation can then be easily automatized by examining specific changes in the feature values. In this paper, we propose the following rules for the automatic segmentation of ERUs, IUs, and activities (see \autoref{fig:segmentation}):
\begin{itemize}
    \item a new ERU begins when at least one element in $X$ changes its value; 
    \item a new IU is initiated when there is a change in $\boldsymbol{x}_\text{c}$. Specifically, if at least one video object starts or stops interacting with others, a new IU arises (
    as in \cite{wachter2013action});
    \item a new activity starts with a specific change in 
    $\boldsymbol{x}_\text{c}$, i.e. when the hand starts interacting with a new object.
\end{itemize}

Let's explain the segmentation strategy with an example. Consider the job of filling a box with five tools. The proposed segmentation would return five similar activities characterized by the hand interacting with a specific object and would involve picking up a tool and placing it inside the box. Each activity would be, in turn, composed of the following four IUs: 
\begin{inparaenum}[(i)]
    \item the human hand grasping the tool from the storage area (involving hand-tool and tool-storage area interactions), 
    \item the hand holding the tool in the air far from the storage area (involving hand-tool interaction but no object-object interaction), 
    \item placing the tool in the box (involving hand-tool and tool-box interactions), and 
    \item the tool becoming integral with the box (involving no hand-object interaction but tool-box interaction). 
\end{inparaenum} 
A new activity starts when a new tool is grasped, and the four IUs are repeated. 

\subsection{Similarity Measures and Clustering}
\label{subsec:similarity}

Following the procedure above, motion features $\boldsymbol{x_\text{m}}$ and context features $\boldsymbol{x_\text{c}}$ can be associated with each video frame, enabling the description of each IU both in terms of motion and of the objects involved in the interaction. In this section, we describe the metrics used to find similar motion patterns among IUs, and discern IUs that involve the same objects. 
Moreover, we propose a machine learning approach to automatically group IUs describing similar interactions without previous knowledge about the accomplished activities.

\subsubsection{Similarity between IUs motion patterns}
To measure the similarity between IUs in terms of motion, we utilize the multi-dimensional Dynamic Time Warping (DTW). DTW is a widely-used distance measure for time series data, and it allows us to compare sequences of different lengths by warping and stretching the time axis. The resulting distance reflects the discrepancy between two context-free IUs, i.e., how different two IUs are based on their motion patterns. The lower the distance, the higher the similarity.
While the distance provided by DTW is a promising candidate measure of the similarity between IUs, determining the condition under which two video segments should be considered as instances of the same interaction is not trivial. To overcome this issue, we exploit the K-means clustering algorithm~\cite{arthur2007kmeans}, where a suitable value of $k$ can be deduced with the elbow method applied to the trend of the Within-Cluster Sum of Squares (WCSS) over $k$. 

\subsubsection{Similarity between IUs contexts}
By definition, each IU is characterized by the same video objects in interaction, thus by a constant $\boldsymbol{x_\text{c}}$. Hence, two distinct $\boldsymbol{x_\text{c}}$ are representative of different IUs. This means that two IUs context are similar (actually identical) if their distance is smaller than 1 for each possible distance metric.
Therefore, to cluster all the IUs involving the same video objects' interactions, we utilize DBSCAN~\cite{schubert2017dbscan} with Euclidean distance and $\epsilon=1$ as the maximum cluster distance.

By combining the two clusterings, we can detect IUs that present similarities both in terms of motion and context.

\subsection{Anomaly Detection in Activity Execution} 

\begin{figure}[!t]
    \centering
	\includegraphics[trim=5cm 2cm 5cm 0.7cm,clip,width=0.95\linewidth]{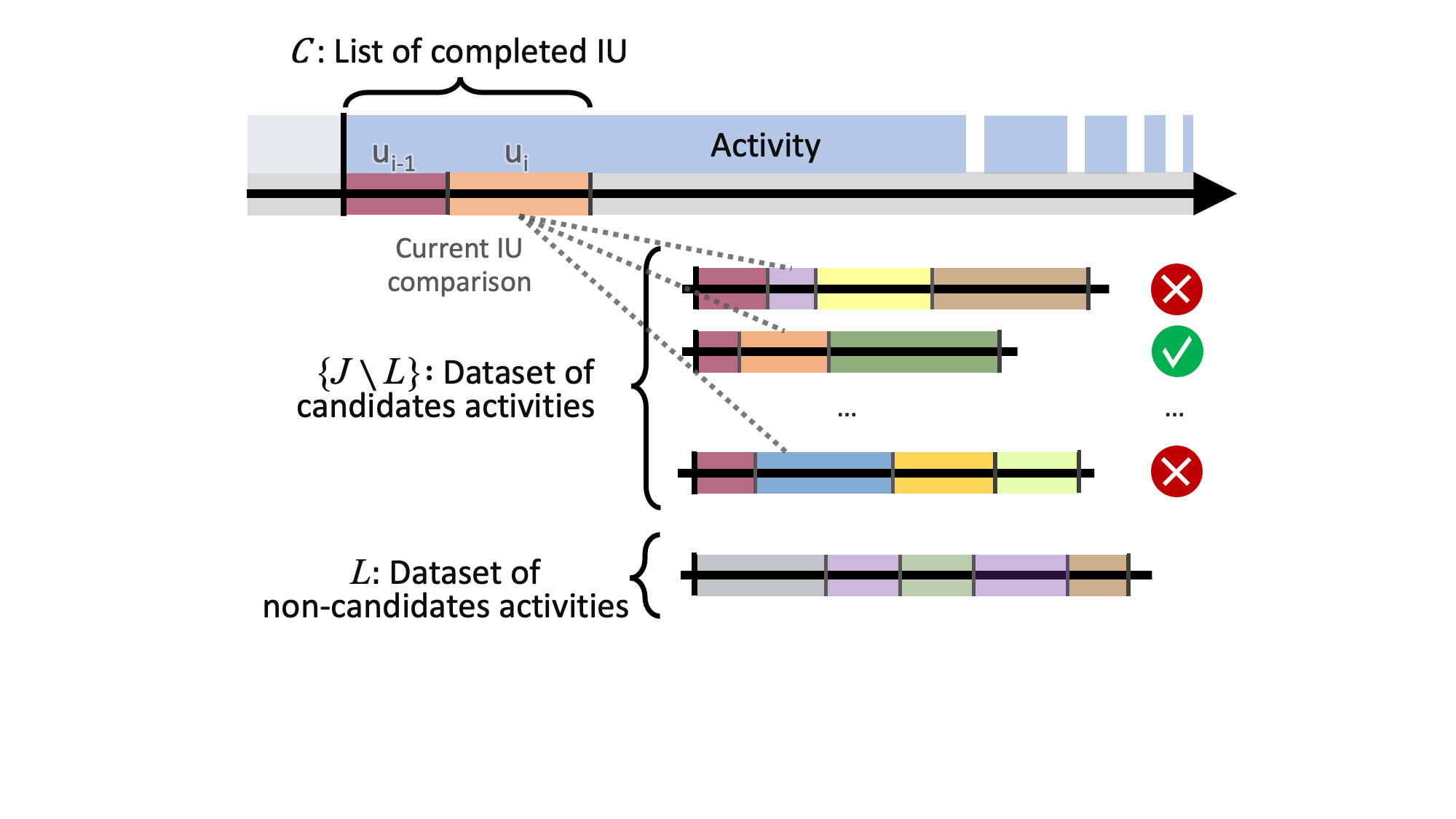}
    \vspace{-1cm}
    \caption{Selection of candidates in anomaly detection algorithm. The first completed IU was compared with the first IUs of all the candidate activities in $J$. This comparison resulted in a division between candidate $\{J \smallsetminus L\}$ and non-candidate $L$ activities. Second completed IU is compared with the second IU of the remaining candidate activities. Once each IU is completed, it is added to the list $C$. This process continues until the end of the activity, when the list $C$ is checked if it perfectly matches with an activity in $\{J \smallsetminus L\}$.}
	\label{fig:failure_detection}
	\vspace{-0.5cm}
\end{figure}

The proposed method to segment and distinguish activities can be particularly useful in applications where the ability to identify in-progress execution deviations is critical. For instance in the anomaly detection in human job executions, a prompt anomaly identification could trigger alerts or corrective actions.

We assume that an activity is correctly performed when all IUs are properly executed and in the correct time order.
The algorithm (see Algorithm \ref{algo:failure_detection}) capitalizes on three lists $J$, $L$, and $C$. The first one contains all the nominal activities required to complete the selected job. The second one comprehends the activities discarded as non-candidates. The latter instead includes the completed IUs. $L$ and $C$ are initially empty.
During the job execution, the recorded scenes are encoded and segmented using the above-mentioned procedure. 
Once a new IU $u_\text{i+1}$ starts, the last completed one, $u_\text{i}$, is added to list $C$ and is compared to all the $i$-$th$ IUs of each activity in the list $\{J \smallsetminus L\}$ in terms of context and motion. 
First, it is checked if $u_\text{i}$ and the candidate $^\text{a}u_\text{i}$ have identical context (i.e., $\boldsymbol{x}_\text{c,i}$==$^\text{a}\boldsymbol{x}_\text{c,i}$). Then, the DTW algorithm is used for the comparison of the motion features ($DTW(\boldsymbol{x}_\text{m,i}, ^\text{a}\boldsymbol{x}_\text{m,i})$). If $d > d_{\text{th,i}}$, that activity is added to the list $L$, which means that it is no longer a candidate activity. Otherwise, the activity remains a candidate. 
If the current IU is not included in any of the activities belonging to the list $\{J \smallsetminus L\}$, an alert will be generated to indicate the occurrence of an anomaly.
The process is repeated for each IU until the current activity ends. \autoref{fig:failure_detection} shows graphically the selection of candidates.
Once the activity ends, the algorithm checks if the sequence of the executed IUs, stored in list $C$, corresponds to the series of IUs of one of the candidate activities 
remaining in $\{J \smallsetminus L\}$ (line 28). At this point, all the activities contained in $\{J \smallsetminus L\}$ have the same time-ordered sequence of $i$ IUs.
If a match is found, it means that the activity was executed correctly. Otherwise, the activity was not completed and the algorithm reports the anomaly.

\alglanguage{pseudocode}
\begin{algorithm}[t]
\small
\caption{Anomaly Detection in Activity Execution}
\label{algo:failure_detection}
\begin{algorithmic}[1]
\linenumbers 
\State $J \gets$ list of the nominal activities of a job
\State $\textbf{d}_{\text{th}} \gets$ distance thresholds for each IU
\State $C \gets \{\}$ \small \Comment{\emph{List of completed IUs}}
\State $L \gets \{\}$ \small \Comment{\emph{List of non-candidate activities}}
\State $i \gets 1$
\While{job is not finished}
    \State \text{activity\_in\_J} $\gets$ \text{True}
    \While{activity is not finished}
        \If{$u_\text{i}$ is finished}
            \State $u_\text{i} \gets$ just completed IU
            \State $\boldsymbol{x}_\text{m,i}$, $\boldsymbol{x}_\text{c,i} \gets$ just completed IU features
            \State Add $u_\text{i}$ to $C$
            \For{\textbf{each} activity \text{a} \textbf{in} $\{J \smallsetminus L\}$}\\
            \small \Comment{\emph{Computing similarity with reference IU}}
                \State $^\text{a}\boldsymbol{x}_\text{m,i}$, $^\text{a}\boldsymbol{x}_\text{c,i} \gets$ $i$-th IU of \text{a} features
                \If{$\boldsymbol{x}_\text{c,i}$ == $^\text{a}\boldsymbol{x}_\text{c,i}$}
                    \State $d \gets \text{DTW}(\boldsymbol{x}_\text{m,i}, ^\text{a}\boldsymbol{x}_\text{m,i})$ 
                    \If{$d > d_{\text{th,i}}$}
                        \small \Comment{\emph{Non-similar motion}}
                        \State Add \text{a} to $L$
                        \small \Comment{a \emph{is non-candidate activity}}
                    \EndIf
                    \Else\small \Comment{\emph{Non-similar context}}
                    \State Add \text{a} to $L$ 
                    \small \Comment{a \emph{is non-candidate activity}}
                \EndIf
            \EndFor
            \If{$\{J \smallsetminus L\} == \emptyset$}
                \State \text{activity\_in\_J} $\gets$ \text{False} \small \Comment{\emph{No such activity in set J}}
            \EndIf
            \State $i \gets i + 1$
        \EndIf
    \EndWhile

    \State \text{activity\_is\_correct} $\gets$ \text{False}
    \If{\text{activity\_in\_J}}
        \For{$\text{a}^*$ \textbf{in} $\{J \smallsetminus L\}$}
            \If{$\text{a}^* \equiv C$} \small \Comment{\emph{Activity executed correctly}}
                \State $L \gets \{\}$,  $C \gets \{\}$ 
                \State \text{activity\_is\_correct} $\gets$ \text{True}
                \State \textbf{break} 
            \EndIf
        \EndFor 
        \If{\textbf{not} \text{activity\_is\_correct}}
            \small \Comment{\emph{Activity not completed}}
        \EndIf 
    \EndIf
\EndWhile
\end{algorithmic}
\end{algorithm}
\section{EXPERIMENTS}
\label{sec:exp}
The evaluation of the method consisted of two experiments. In the first one, we validated the proposed video segmentation and the IUs similarity recognition, while in the second experiment, we tested our anomaly detection algorithm on the activity execution. The experimental setup envisioned an RGB camera (Intel RealSense D435i) mounted in top shot (bird's eye) view, and the image plane was aligned with the working plane (i.e., the tabletop of the workbench).

Since object detection and hand pose estimation are extensively studied and beyond the scope of this paper, in our experiments, we have opted for a reliable marker-based object detection system and an open-source hand detector.
Specifically, we exploit ArUco markers as a 3D object detection method. However, the same method did not give satisfactory results in hand detection due to the different configurations that the hand can have during the manipulation task, eventually hiding the marker, and since the marker hindered the natural movements of participants. For this reason, we utilized the MediaPipe Hand Detector, which features 21 hand landmarks. For simplicity, we detected only the right hand and evaluated the method in 2D (in the image plane, which is aligned with the workbench tabletop). As a result, feature $\varphi$ was not taken into account.
The architecture has been developed in Python, on Ubuntu 20.04 and ROS Noetic, exploiting the \href{https://dtaidistance.readthedocs.io/en/latest/index.html}{DTAIdistance} library~\cite{meert2020dtaidistance}  for the multi-dimensional DTW and the k-means clustering and \href{https://scikit-learn.org/stable/index.html}{Scikit-Learn} library~\cite{pedregosa2011scikit-learn} for the DBSCAN with Euclidean distance.

\subsection{Activity Recognition Validation (Exp 1)}
\label{ssec:exp1}

\begin{figure} [t]
    \centering
    \includegraphics[scale=0.25]{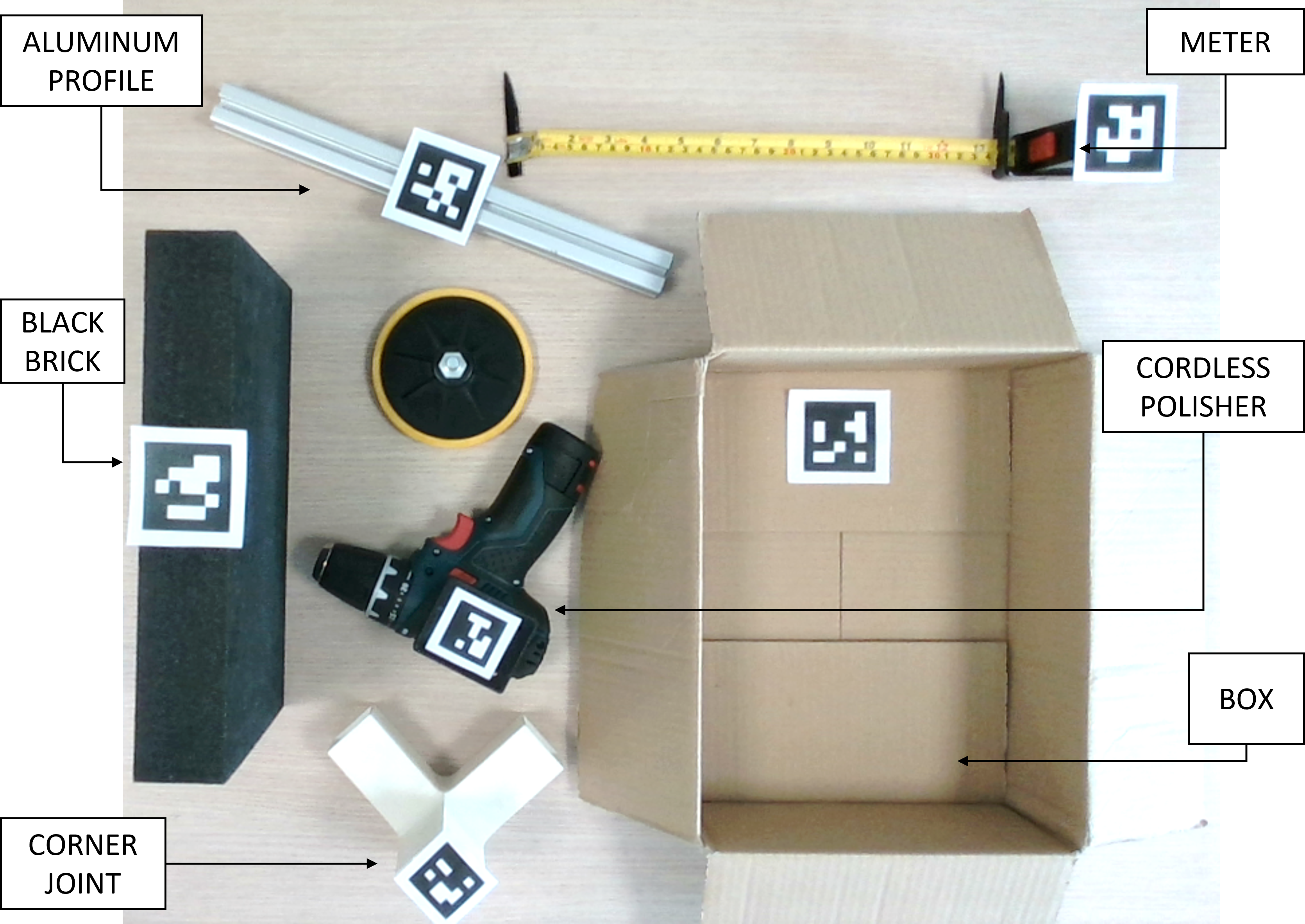}
    \caption{Objects used in the experiments.}
	\label{fig:objects}
	\vspace{-0.5cm}
\end{figure}

\begin{figure} [t]
    \centering 
    \includegraphics[trim=5cm 1.5cm 5cm 0.6cm, clip, width=0.9\linewidth]{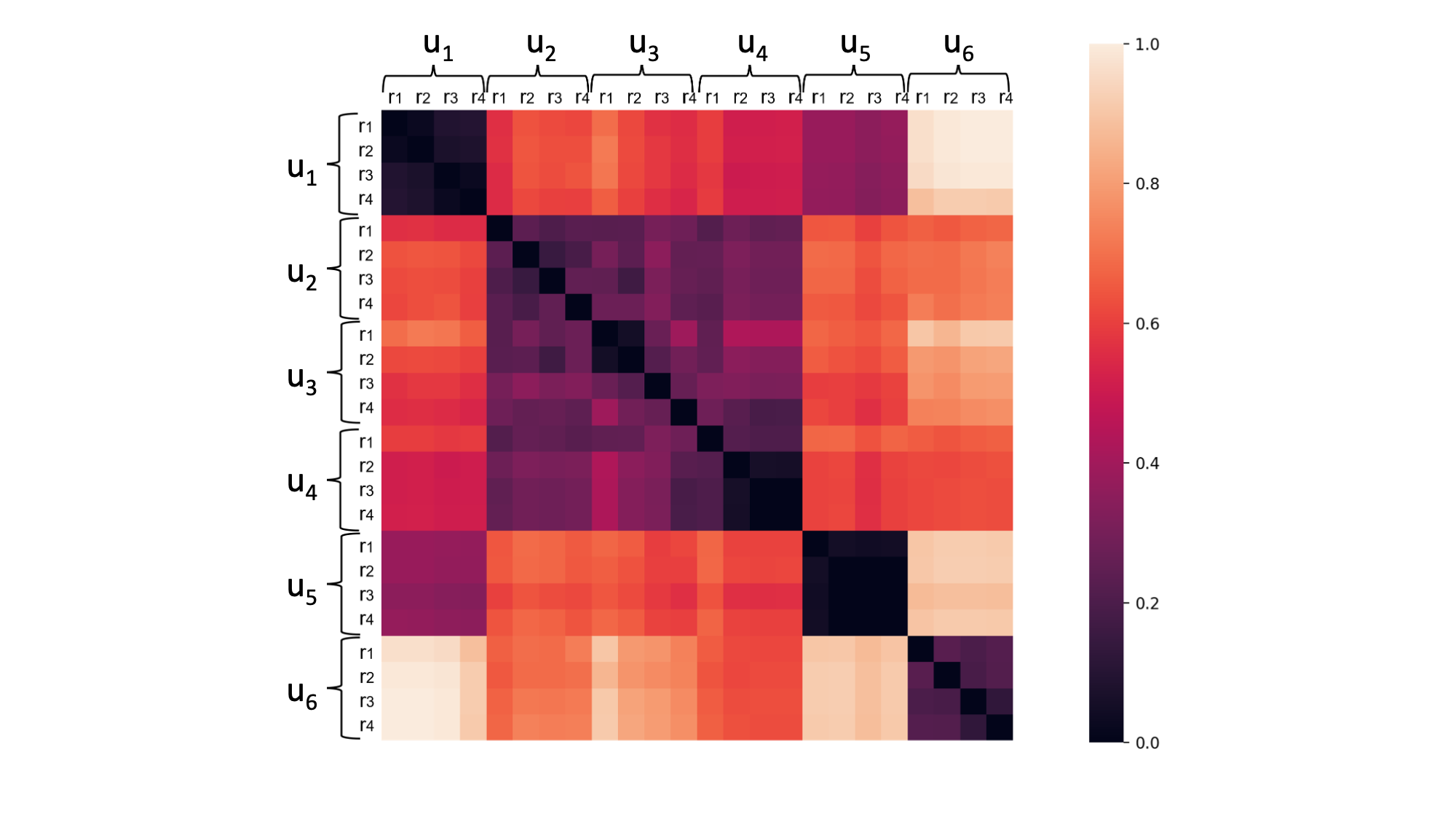}
    \vspace{-0.1cm}
    \caption{Single subject Confidence Matrix resulting from the comparison of context-free IUs. The IUs are respectively: $u_{\text{1}}$ grasp the profile, $u_{\text{2}}$ put the profile in the box, $u_{\text{3}}$ measure the profile, $u_{\text{4}}$ assemble the profile and corner, $u_{\text{5}}$ disassemble the profile and corner, $u_{\text{6}}$ polish the black brick surface. $r_{\text{i}}$ corresponds to the $i$-$th$ repetition. A darker color in the matrix indicates a higher similarity between the motion patterns of the IUs being compared.}
	\label{fig:subject1}
	\vspace{-0.5cm}
\end{figure}

In the first experiment, we asked $N_{\text{sub}} = 10$ subjects, 7 males and 3 females ($25.6 \pm 1.2$ years old), to perform a set of five activities for $N_{\text{rep}} = 4$ repetitions. These activities involved the following objects: an aluminum profile, a corner joint, a meter, a box, a polisher, and a black brick (shown in \autoref{fig:objects}).
After the completion of each activity, the associated features $X$, normalized in $[0, 1]$, were filtered using the opening-closing filter \cite{bhutada2022opening}, then the segmentation was performed. This implies that a new IU is generated only if a particular interaction persists for a consecutive number of frames.
The activities and IUs segmented by the algorithm are listed below (the semantic labels are not used by the algorithm and are reported only to clarify the experiment): 
\begin{enumerate}[label=\Roman*)]
    \item \textit{Boxing}: grasp the profile, put the profile inside the box, leave the profile inside the box;
    \item \textit{Measuring}: grasp the profile, put the profile close to the meter, leave the profile close to the meter;
    \item \textit{Assembly}: grasp the profile, interlock the profile and the corner joint, leave the complete assembly;
    \item \textit{Disassembly}: profile and corner joint already interlocked, disassemble the profile and the corner joint, move away from the corner while holding the profile;
    \item \textit{Polishing}: grasp the polisher, polish the black brick with back and forth motions, move away from the black brick surface while holding the polisher.
\end{enumerate}
To further evaluate the similarity between IUs, we selected a subset of size $N_{\text{IU}} = 6$ of representative IUs: 
\begin{enumerate}
    \item grasp the profile, from \textit{measuring}; 
    \item put the profile inside the box, from \textit{boxing};
    \item put the profile close to the meter, from \textit{measuring};
    \item interlock the profile and the corner joint, from \textit{assembly};
    \item disassemble the profile and the corner joint, from \textit{disassembly};
    \item polish the black brick with back and forth motions, from \textit{polishing}.
\end{enumerate}
We initially analyzed the IUs extracted from the $N_{\text{rep}}=4$ repetitions of each activity by a single subject, computing similarities in terms of motion features and context features.
The similarities between the context-free IUs were evaluated using DTW and a confidence matrix of size $(N_{\text{IU}} \times N_{\text{rep}})x(N_{\text{IU} } \times N_{\text{rep}})$ was generated to show the distances between each couple of IUs.
\autoref{fig:subject1} shows the Single subject Confidence Matrix (SCM) for subject 5, which reports similarities between context-free IUs. The distances obtained from DTW were normalized to a range between $0$ and $1$, and dark patches indicate lower distance, hence higher similarity between IUs. 

\begin{figure} [t]
    \centering
    \includegraphics[trim=5cm 1.5cm 5cm 0.6cm, clip, width=0.9\linewidth]{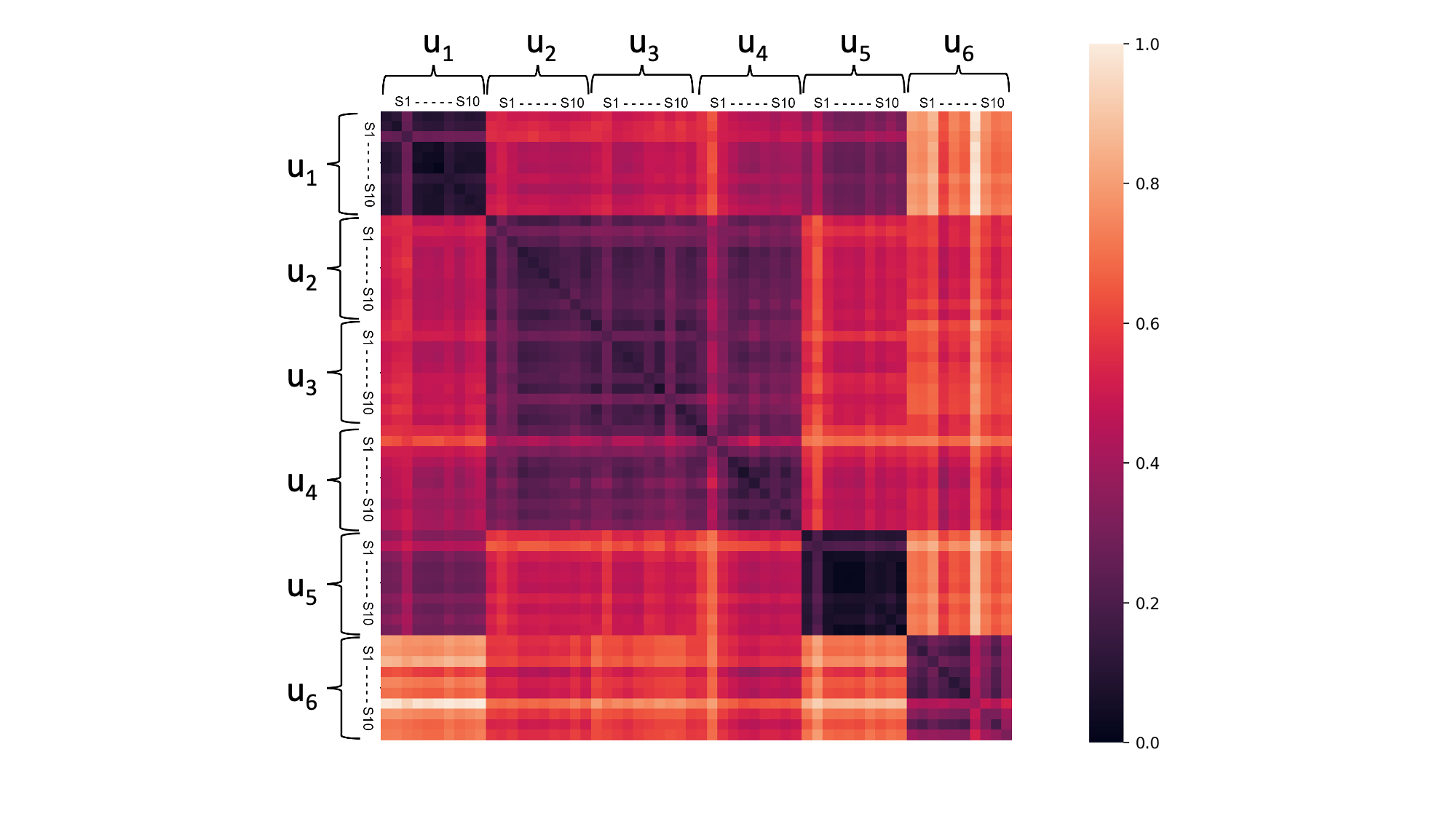}
    \caption{Multi-subject Confidence Matrix resulting from the comparison of context-free IUs. The IUs are respectively: $u_{\text{1}}$ grasp the profile, $u_{\text{2}}$ put the profile in the box, $u_{\text{3}}$ measure the profile, $u_{\text{4}}$ assemble the profile and corner, $u_{\text{5}}$ disassemble the profile and corner, $u_{\text{6}}$ polish the black brick surface. $s_{\text{1}} \dots s_{\text{10}}$ corresponds to the subject number, which ranges from 1 to 10. A darker color in the matrix indicates a higher similarity between the motion patterns of the IUs being compared.}
    \label{fig:multi_sub_matrices}
    \vspace{-0.2cm}
\end{figure}

To evaluate instead the similarity of the same IU performed by different subjects, we conducted a multi-subject analysis by computing the distances between all the repetitions across all the participants. We filled the corresponding confidence matrix Multi subjects Confidence Matrix (MCM) with size $(N_{\text{IU}} \times N_{\text{sub}})x(N_{\text{IU}} \times N_{\text{sub}})$ (see \autoref{fig:multi_sub_matrices}).
Each element is computed as:
\begin{equation*} \begin{split}
MCM(\text{a},\text{x},\text{b},\text{y}) = \frac{1}{N_{\text{rep}}^2} \sum\limits_{m=1}^{N_{\text{rep}}} \sum\limits_{n=1}^{N_{\text{rep}}} DTW((u_\text{a}, s_\text{x}, r_\text{m}), (u_\text{b}, s_\text{y}, r_\text{n}))
\end{split}
\end{equation*}
where $\text{a,b} \in [1, N_{\text{IU}}]$, $\text{x,y} \in [1,N_{\text{sub}}]$. $u_\text{a}$ and $u_\text{b}$ identify the couple of IUs we are comparing, while $s_\text{x}$ and $s_\text{y}$ denote the subjects. In other words, each element of MCM represents the distance between the average performance of all the $N_{\text{rep}}$ repetitions of IU $u_\text{a}$ for subject $s_\text{x}$ and the average performance of all the $N_{\text{rep}}$ repetitions of IU $u_\text{b}$ for subject $s_\text{y}$. Distances were then normalized in $[0, 1]$.

\begin{figure} [t]
    \centering
    \includegraphics[trim=0.4cm 0.9cm 0.2cm 0.9cm, clip,width=\linewidth]{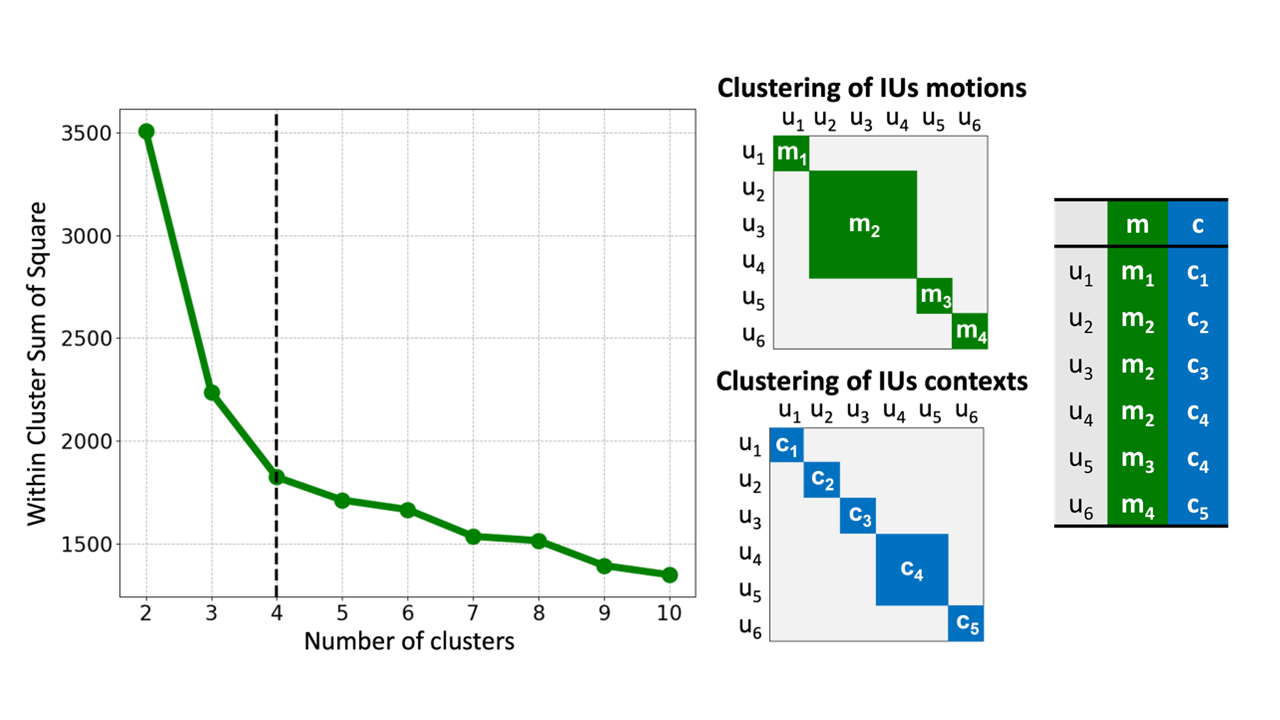}
    \caption{Results of clustering IUs from multi-subject experiment. K-means clustering is used for clustering IUs motion patterns. The best $k$ value $k=4$ is found with the elbow method (left) and we obtain clusters $m_\text{1}$, $\dots$, $m_\text{4}$. Results show $u_{\text{2}}$, $u_{\text{3}}$, and $u_{\text{4}}$ as a unique cluster (top-center). DBSCAN is used for clustering IUs contexts and it results in 5 clusters ($c_\text{1}$, $\dots$, $c_\text{5}$), one for each IU except for $u_{\text{4}}$ and $u_{\text{5}}$, which are grouped together in a single cluster (bottom-center). Then the two clustering results are merged to obtain a unique combination of motion and context features ($m_\text{i}$, $c_\text{i}$) for each IU (right). }
\label{fig:multi_sub_clustering}
    \vspace{-0.6cm}
\end{figure}

We further analyzed the results by clustering all executions of the IUs by all subjects using k-means with DTW as distance metric.
To determine the optimal number of clusters, we ran the algorithm $10$ times 
for $k \in [1,10]$ and selected the minimum WCSS for each $k$. 
The best value of $k=4$ was given by the elbow method (see \autoref{fig:multi_sub_clustering} (left)). In particular, IUs of the \textit{assembly}, \textit{measuring}, and \textit{boxing} activities were grouped together in the same cluster (see \autoref{fig:multi_sub_clustering} (top-center)). 
At the same time, to compare IUs context, we used the DBSCAN algorithm with Euclidean distance and maximum cluster distance $\epsilon = 1$. In this case, the clustering algorithm identified a total of $k=5$ clusters, one for each type of IU except for those from \textit{assembly} and \textit{disassembly} activities, which were grouped into a single cluster (see \autoref{fig:multi_sub_clustering} (bottom-center)). To obtain a unified combination of motion and context features characterizing each IU, the two clustering results are combined together as shown in \autoref{fig:multi_sub_clustering} (right).
 
\begin{figure} [t]
    \centering
	\includegraphics[trim=5cm 2cm 4.2cm 0.6cm, clip, width=0.9\linewidth]{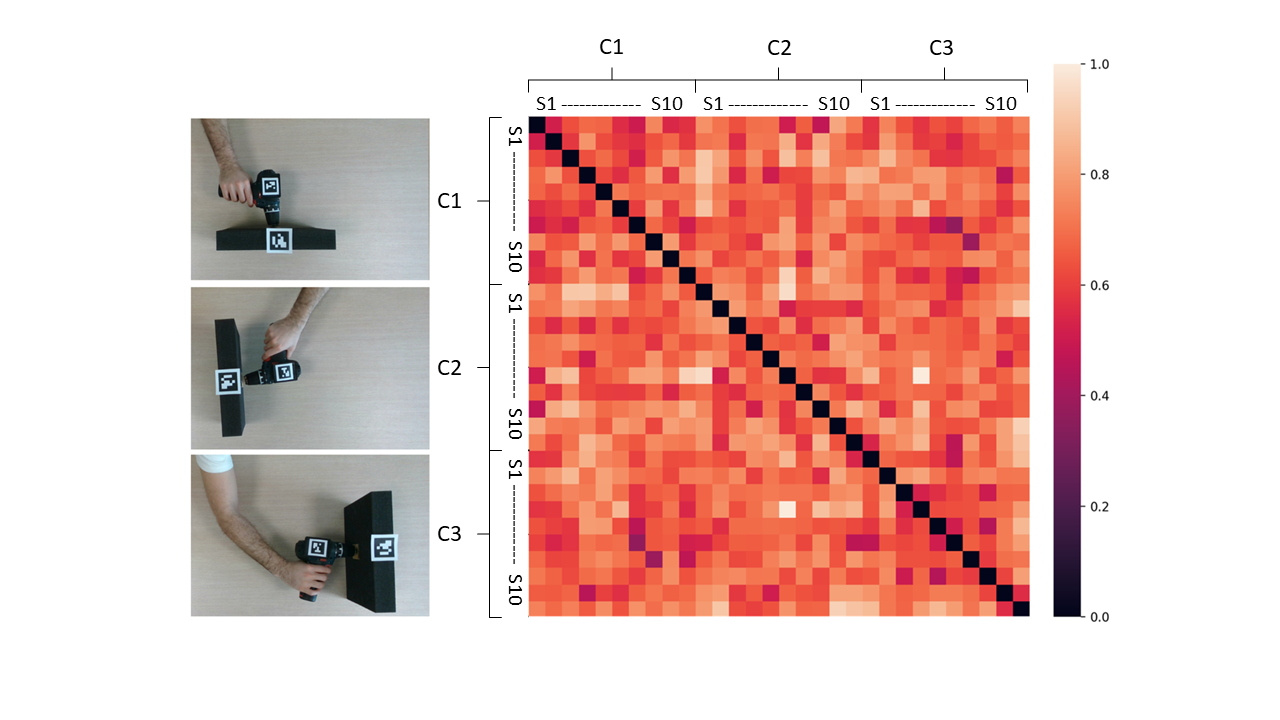}
	\vspace{-0.2cm}
    \caption{Drilling activity being performed in three different configurations labeled as C1, C2, and C3 (left). Confidence matrix obtained by calculating the distance between all the IUs from each configuration (right). $s_{\text{1}} \dots s_{\text{10}}$ corresponds to the subject number, which ranges from 1 to 10.}
	\label{fig:drilling_conf_m}
	\vspace{-0.5cm}
\end{figure}

Besides, we conducted an additional analysis to verify the robustness of the detection with respect to variations of the absolute poses of the involved video objects within the same IUs. 
In particular, we asked the same participants to perform a \textit{drilling} activity where the IUs included: 
\begin{inparaenum}
    \item grasp the drill, 
    \item drill the black brick for 5 seconds, and 
    \item move away from the surface while holding the drill. 
\end{inparaenum}
This activity was carried out while changing the absolute position of the drill and the black brick in three different configurations (C1, C2, C3), as illustrated in \autoref{fig:drilling_conf_m} (left), for a total of 30 activities executed. In \autoref{fig:drilling_conf_m} (right) we reported the confidence matrix of the second IU.

\subsection{Anomaly Detection Algorithm Validation (Exp 2)}
\label{ssec:exp2}
Our second experiment focused on identifying anomalies in a job that consisted of two distinct activities:
\begin{inparaenum}[I)]
    \item polishing the black brick surface, and 
    \item measuring its thickness. 
\end{inparaenum}
Activity I) consisted of three IUs: 
\begin{inparaenum}
    \item grasp the polisher;
    \item polish the brick surface employing back and forth motions;
    \item move away from the surface and release the polisher.
\end{inparaenum}
Activity II) involved two IUs:
\begin{inparaenum}
    \item grasp the brick;
    \item place the brick close to the meter.
\end{inparaenum}
The aim of this evaluation is to demonstrate a potential application that utilizes the ability to distinguish between similar and non-similar interactions. While the algorithm is designed for operating online, its assessment is conducted offline and no computational time-related performance metrics are reported in this study. 

We asked $N_{\text{sub}} = 7$ subjects, $6$ males and $1$ female ($25.7 \pm 1.1$ years old) to perform the job correctly $3$ times. Using the procedure described in \autoref{subsec:segmentation}, we automatically segmented each of the filtered $N_{\text{job\_correct}} = 21$ job executions into $N_{\text{activity}} = 2$ and $N_{\text{IU}} = 5$ IUs. 
Subsequently, we asked each subject to repeat 2 flawed job executions ($J_\text{1}$ and $J_\text{2}$), leading to $N_{\text{job\_flawed}} = 14$. 
In $J_\text{1}$, we instructed the subjects to fail the second IU of activity I) by stopping halfway the polishing, while in $J_\text{2}$, we asked them to fail the second IU of activity II) by not measuring the brick.
The accuracy of the anomaly detection algorithm presented in Algorithm \ref{algo:failure_detection} was evaluated through a cross-validation repeated for $N_{\text{iterations}} = 10$ rounds. In each round, we randomly divided the set of correct executions, which contains $N_{\text{job\_correct}}$ samples, into a training set and a test set. The training set contained $N_{\text{job\_training}} = 14$ randomly selected samples, while the test set contained the remaining correct samples plus all executions of $J_\text{1}$ and $J_\text{2}$, i.e., $N_{\text{job\_test}} = N_{\text{job\_correct}} + N_{\text{job\_flawed}} - N_{\text{job\_training}} = 21$. 
The algorithm requires a nominal execution of the job and the distance threshold vector $\boldsymbol{d}_{\text{th}}$.
The first was obtained by calculating the barycenter $b_{\text{u}_\text{i}}$ of each IU cluster (containing  $N_{\text{job\_training}}$ samples), $i \in [1, N_{\text{IU}}]$. While each element of $\boldsymbol{d}_{\text{th}}$ corresponds to $d_{\text{th,i}} = \mu_{\text{u}_\text{i}} + 2\sigma_{\text{u}_\text{i}}$, where $\mu_{\text{u}_\text{i}}$ and $\sigma_{\text{u}_\text{i}}$ are mean and standard deviation of DTW distance on the motion features of each IU from its corresponding barycenter.  
Finally, the nominal activities generated by the sequence of $b_{\text{u}_\text{i}}$ are stored in $J$.
At each iteration, the selected $N_{\text{job\_training}}$ were used for retrieving $b_{\text{u}_\text{i}}$ and $d_{\text{th,i}}$ for each IU. 
The confusion matrix in \autoref{tab:failure_det_confusion_m} shows the accuracy in recognizing correct and warped executions in terms of IU and activities.
Note that the total number of IUs and activities is obtained using the following formulas: $N_{\text{tot\_IU}} = N_{\text{job\_test}} \times N_{\text{iterations}} \times N_{\text{IU}}$, and $N_{\text{tot\_activity}} = N_{\text{job\_test}} \times N_{\text{iterations}} \times N_{\text{activity}}$.

\begin{table}[!t]
\caption{\small Confusion matrix of Anomaly Detection}
    \centering
\vspace{-2mm}
\begin{tabular}{lc|cc}
        \multicolumn{2}{l}{\textit{IU level}} &  \multicolumn{2}{c}{\textbf{Predicted}} \\
        & & Negative & Positive  \\
    \hline
        \multirow{2}{*}{\textbf{Actual}} & Negative  & 858 & 52 \\ 
        & Positive  & 0 & 140 \\ 
    \hline
    & \multicolumn{3}{c}{Accuracy = $95.0\% \pm 2.7\%$} \\
    \label{tab:failure_det_confusion_m}
\end{tabular}
\begin{tabular}{lc|cc}
        \multicolumn{2}{l}{\textit{Activity level}} &  \multicolumn{2}{c}{\textbf{Predicted}} \\
        & & Negative & Positive  \\
    \hline
        \multirow{2}{*}{\textbf{Actual}} & Negative  & 252 & 44 \\ 
        & Positive  & 0 & 124 \\ 
    \hline
    & \multicolumn{3}{c}{Accuracy = $89.5\% \pm 5.1\%$} \\
\end{tabular}
\vspace{-0.6cm}
\end{table}
\section{DISCUSSION AND CONCLUSIONS}
\label{sec:disc_conc}

In this paper, we proposed a bottom-up approach for recognizing activities by analyzing object-object and hand-object interactions in terms of motion and context information. 
Experiments in \autoref{ssec:exp1} demonstrated the capability of the framework to identify and group similar context-free interactions.
This indicates that our scene encoding and features-based representation succeeded in comprehensively describing manual activities and identifying video objects interaction' changes during the execution. 
Strong points of our encoding include the automatic segmentation of activities and IUs, as well as the possibility of separating contextual information from motion information. Considering the motion features exclusively, we can recognize IUs that involve similar motion patterns. Within our experiments, DTW identifies similarities between \textit{boxing}, \textit{measuring}, and \textit{assembly} (see SCM in \autoref{fig:subject1}). This outcome is motivated by the fact that the IUs mentioned above share  
\begin{inparaenum}[(i)]
    \item the hand holding of the profile, 
    \item the approaching of a tool (i.e. box, meter, or corner joint),
    \item the hand release of the profile, and 
    \item the hand moving away.
\end{inparaenum}
As shown in \autoref{fig:multi_sub_matrices}, the IUs were grouped in an analogous way to SCM, no matter who executed the activity. We can deduce that our framework is robust to the variability induced by different subjects in the activity execution. Moreover, the algorithm was robust to changes in the absolute poses of the video objects (see \autoref{fig:drilling_conf_m}). 
Additionally, our method shows promising results in leveraging the recognition of both type similar and non-type similar interactions to ensure job performance without anomalies. A preliminary experiment indicated a high success rate and consistent results across iterations using different references (\autoref{tab:failure_det_confusion_m}). Interestingly, the errors committed were only false positive, meaning that the algorithm occasionally failed in detecting correctly performed IUs. This was probably due to the limited size of the training set, which may not capture the full range of variations in correct activity executions. 

However, we acknowledge several limitations of our approach, including heavy reliance on accurate object and hand detection, the lack of occlusion handling, and the evaluation in 2D with a fixed camera orientation. Moreover, the current version of our method only describes interactions for each hand separately and does not consider any link between activities performed by the two hands, even when they occur simultaneously.
In future works, we plan to address these limitations and further refine our method to enhance its capabilities and relevance in robotics applications.

\bibliographystyle{IEEEtran}
\bibliography{lib}

\end{document}